%% file: main.tex
\documentclass{article}

\usepackage{url}
\usepackage{cite}
\usepackage{soul}
\usepackage{ulem}
\usepackage{arxiv}
\usepackage{xspace}
\usepackage{siunitx}
\usepackage{graphicx}
\usepackage{textcomp}
\usepackage{graphicx}
\usepackage{multirow}
\usepackage{booktabs}
\usepackage{placeins}
\usepackage{makecell}
\usepackage{graphicx}
\usepackage{multicol}
\usepackage{adjustbox}
\usepackage{subcaption}
\usepackage{algorithmic}
\usepackage[table,xcdraw]{xcolor}
\usepackage{amsmath,amssymb,amsfonts}
\usepackage[pagebackref,breaklinks,colorlinks]{hyperref}

\def\BibTeX{{\rm B\kern-.05em{\sc i\kern-.025em b}\kern-.08em
    T\kern-.1667em\lower.7ex\hbox{E}\kern-.125emX}}
\title{Memory Efficient Continual Learning for Edge-Based Visual Anomaly Detection}

\author{
    Manuel Barusco \\ 
    University of Padova, Italy \\ 
    \texttt{manuel.barusco@phd.unipd.it} \\ \And 
    Lorenzo D'Antoni \\ 
    University of Padova, Italy \\ 
    \texttt{lorenzo.dantoni@studenti.unipd.it} \\ \And
    Davide Dalle Pezze \\ 
    University of Padova, Italy \\ 
    \texttt{davide.dallepezze@unipd.it} \\ \And
    Francesco Borsatti \\ 
    University of Padova, Italy \\ 
    \texttt{francesco.borsatti.1@phd.unipd.it} \\ \And
    Gian Antonio Susto \\ 
    University of Padova, Italy \\ 
    \texttt{gianantonio.susto@unipd.it} \\
}

\begin{document}

\maketitle

\input{sec/0_abstract}    
\input{sec/1_intro}
\input{sec/2_related}
\input{sec/3_methodology}
\input{sec/4_expsettings}

\input{sec/5_results}

\input{sec/6_conclusions}

\bibliographystyle{IEEEtran}
\bibliography{main}

\end{document}

%% file: sec/0_abstract.tex
\begin{abstract}
Visual Anomaly Detection (VAD) is a critical task in computer vision with numerous real-world applications.
However, deploying these models on edge devices presents significant challenges, such as constrained computational and memory resources.
Additionally, dynamic data distributions in real-world settings necessitate continuous model adaptation, further complicating deployment under limited resources.
To address these challenges, we present a novel investigation into the problem of Continual Learning for Visual Anomaly Detection (CLAD) on edge devices. 
We evaluate the STFPM approach, given its low memory footprint on edge devices, which demonstrates good performance when combined with the Replay approach.
Furthermore, we propose to study the behavior of a recently proposed approach, PaSTe, specifically designed for the edge but not yet explored in the Continual Learning context.
Our results show that PaSTe is not only a lighter version of STPFM, but it also achieves superior anomaly detection performance, improving the f1 pixel performance by 10\% with the Replay technique.
In particular, the structure of PaSTe allows us to test it using a series of Compressed Replay techniques, reducing memory overhead by a maximum of 91.5\% compared to the traditional Replay for STFPM.
Our study proves the feasibility of deploying VAD models that adapt and learn incrementally on CLAD scenarios on resource-constrained edge devices.
\end{abstract}

%% file: sec/1_intro.tex
\section{Introduction}
\label{sec:intro}

\begin{figure*}[!thbp]
  \centering
  \includegraphics[width=0.8\linewidth]{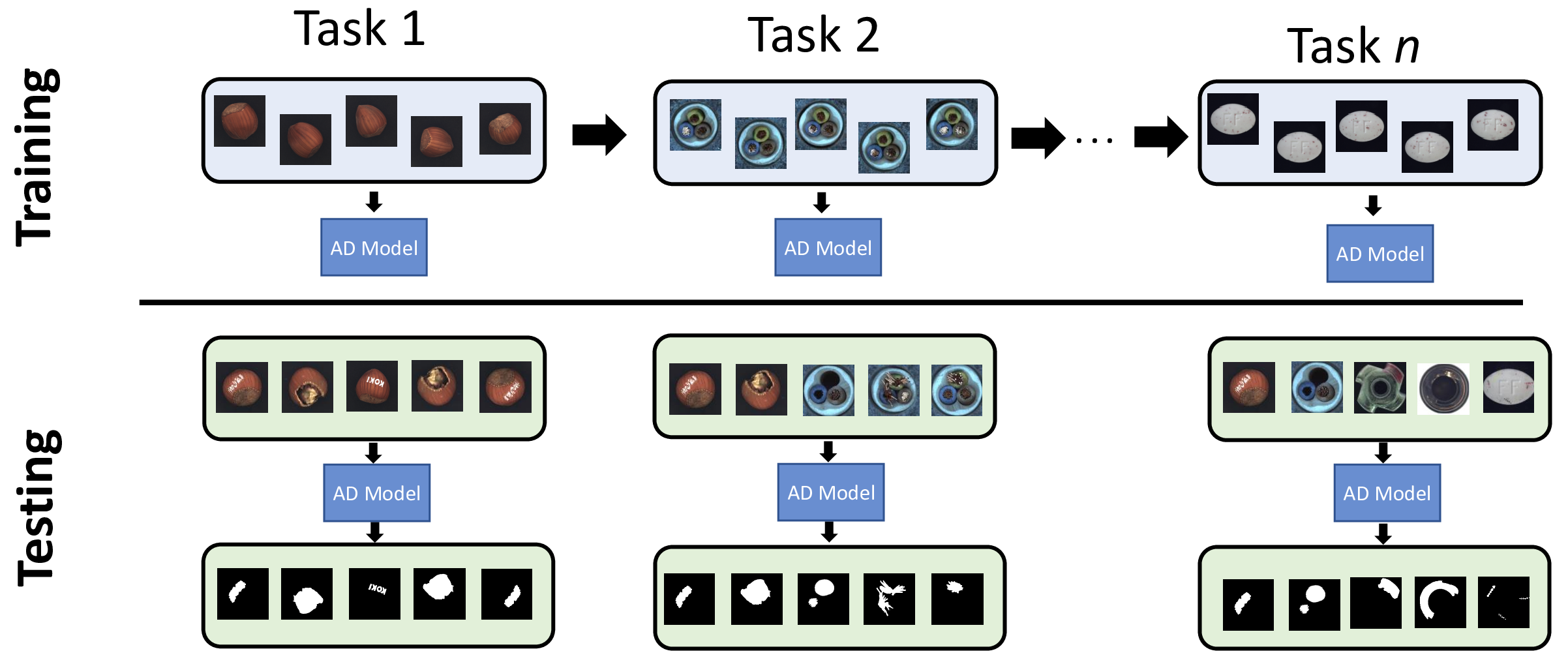}
   \caption{Considered CL setting for the AD problem \cite{clad}. Each task corresponds to a new item. The AD models must be able to detect the anomalous products (image-level) and the defects inside the image (pixel-level) of a new item while remembering to perform well on previously seen items.}
   \label{fig:CLAD_scheme}
\end{figure*}

Visual Anomaly Detection (VAD) is a critical Computer Vision task that involves identifying images containing anomalies and localizing the specific pixels responsible for the irregularity. By leveraging unsupervised learning, VAD avoids the time-consuming and costly process of collecting labeled data for pixel-level anomaly annotation. 
VAD has a significant potential impact across various industries, with applications such as manufacturing, medical imaging, and autonomous vehicles \cite{bao2023bmad} \cite{mvtec} \cite{chan2021segmentmeifyoucan}.

VAD has gained much research attention in recent years, and many VAD methods have been developed, focusing mainly on performance. 
Only in the last period have other challenges been considered in the development of the methods.
For example, \cite{noisy} considers the scenario where training and test data are noisy, with the images not well illuminated or presenting other kinds of noise. 
Other studies considered the challenge of deploying VAD methods in edge scenarios, where resources are limited \cite{paste}, analyzing the compatibility of established VAD methods with edge limitations.

Lastly, \cite{clad,pezze2022continual,xie2024iad} consider the development of VAD methods in the Continual Learning (CL) scenario. 
This last challenge deals with the shifts in the input data distribution that can occur over time in real-world scenarios. For instance, in industrial settings, novel objects are continually introduced (see Fig. \ref{fig:CLAD_scheme}), requiring defect detection, while in medical imaging, newly discovered anatomical structures necessitate analysis for anomaly detection. 

In this work, we want to tackle the unexplored challenge of the development of VAD solutions that can work in an edge scenario under the CL setting. 
As a VAD model with a low memory footprint on edge devices, as demonstrated by previous studies \cite{paste}, we assess the STFPM model within the CLAD setting for edge devices with limited resources.
Furthermore, we propose to study the behavior of a recently proposed approach, Paste, specifically designed for the edge but not yet explored in the Continual Learning context.

Our results show that using the Replay technique, PaSTe model is not only a lighter version of STPFM, but it also achieves superior anomaly detection performance, improving the f1 pixel performance by 10\%.
In particular, the structure of PaSTe allows us to test it using a series of Compressed Replay techniques, reducing memory overhead by a maximum of 91.5\% compared to the traditional Replay for STFPM.

The main contributions of our work can be summarized as follows: 
\begin{enumerate}
    \item We tackled for the first time the challenge of Continual Learning Visual Anomaly Detection (CLAD) in the Edge Setting by testing the STFPM model using the Replay approach.
    \item Moreover, we propose to use for the first time the Paste approach in the Continual Learning scenario, proving that it is able to have better performance than STFPM while reducing several metrics such as training and inference time.
    \item Moreover, given the architecture of Paste, we propose a series of Compressed Replay techniques that allow to reduce further the memory.
\end{enumerate}

In addition, we make the code available for reproducibility \footnote{\url{https://bitbucket.org/papers_vad_group/clad_paste/src/master/}}.

The outline of the paper is as follows: Sec. \ref{sec:related_work} describes the literature of VAD algorithms, with a focus on the VAD methods suitable for edge deployment and for Continual Learning. Sec. \ref{sec:methodology} describes the implementation details of our solution.
Sec. \ref{sec:experimental_setting} shows the experimental settings while Sec. \ref{sec:results} discusses the obtained results in our study.
Eventually, Sec. \ref{sec:conclusion} concludes this work delineating potential future research directions

%% file: sec/2_related.tex
\section{Related Work}
\label{sec:related_work}

Sec. \ref{subsec:related_work_vad} provides a review of the literature on VAD, with a particular focus on approaches designed for edge computing environments. Sec. \ref{subsec:continual_learning} presents a concise overview of the relevant work in Continual Learning.

\begin{figure*}[!th]
  \centering
  \begin{subfigure}[b]{0.49\linewidth}
    \centering
    \includegraphics[width=\linewidth]{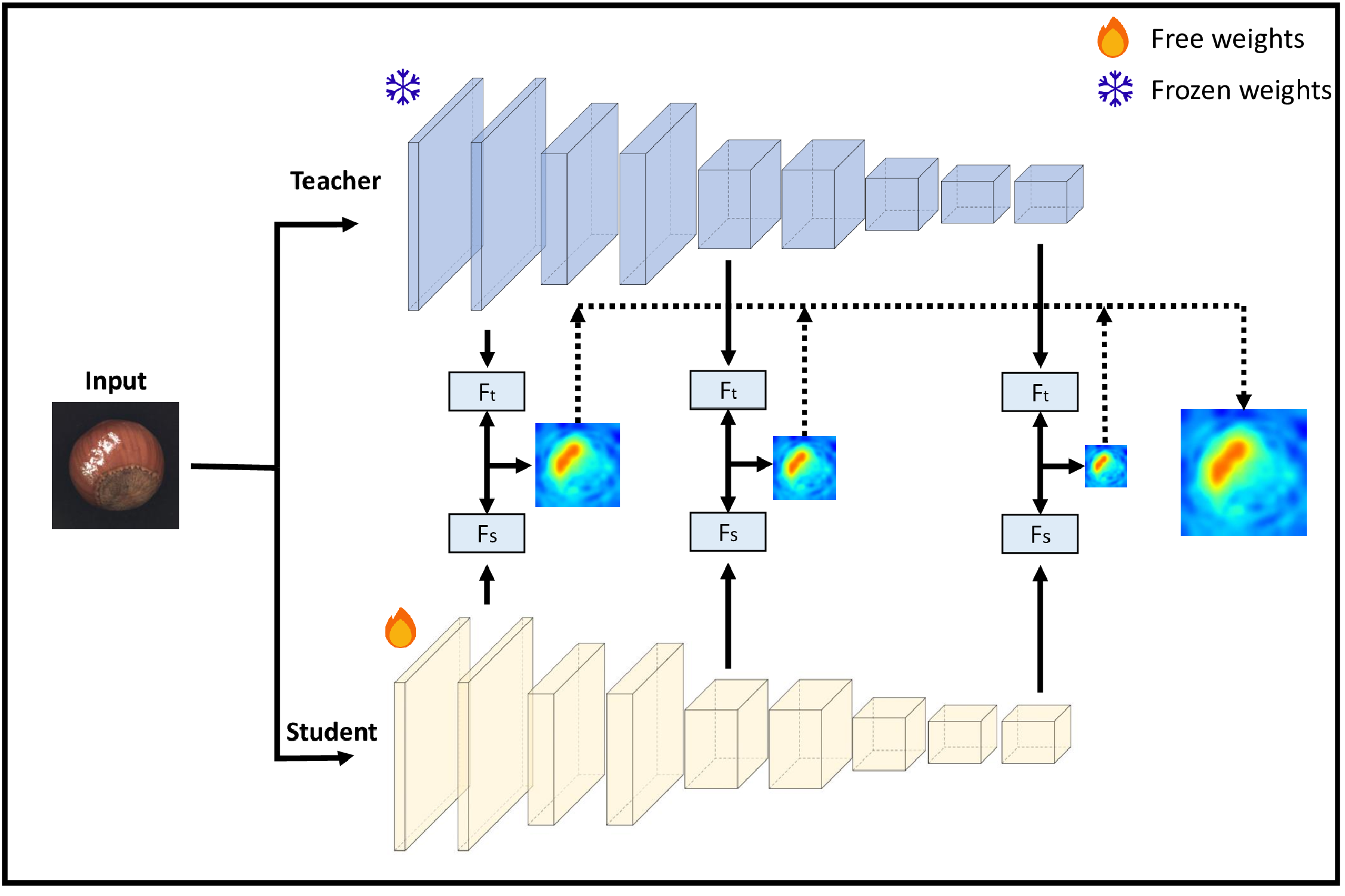}
    \caption{STFPM}
    \label{fig:STFPM_architecture}

  \end{subfigure}
  \hfill
  \begin{subfigure}[b]{0.49\linewidth}
    \centering
    \includegraphics[width=\linewidth]{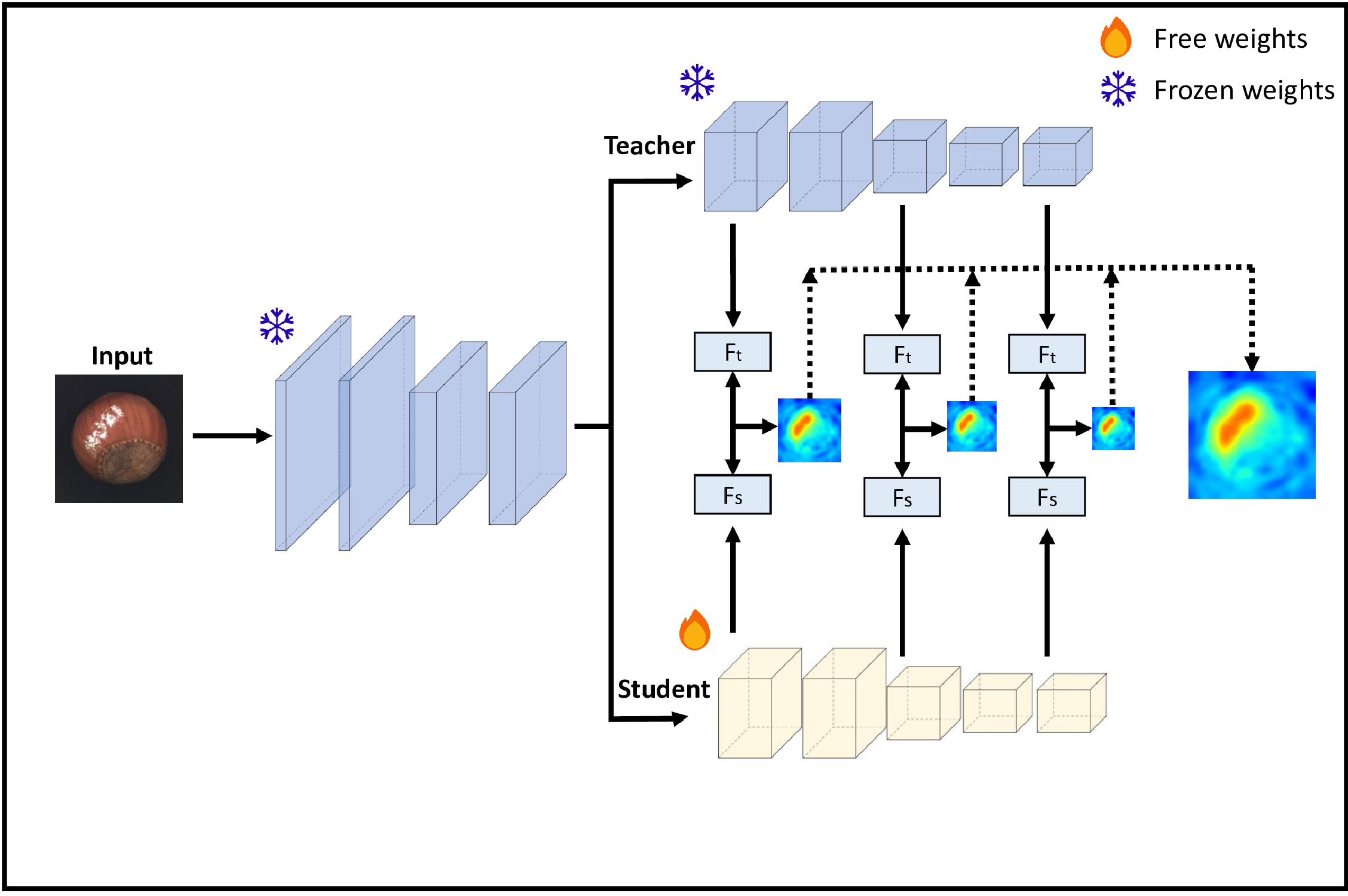}
    \caption{Paste}
    \label{fig:Paste_architecture}
  \end{subfigure}
  \caption{Comparison between the STFPM and PaSTe architectures \cite{paste}. PaSTe reduces the memory required and computation resources to the minimum. }
  \label{fig:OurApproach}
\end{figure*}

\subsection{Visual Anomaly Detection}
\label{subsec:related_work_vad}

Anomaly Detection (AD) has numerous applications in Computer Vision, including manufacturing, healthcare, autonomous vehicles, and security. 
Many AD models provide a binary sample level classification (normal or anomalous) to support decision-making. Still, decision-makers often need more information about why a sample is anomalous. Enhancing model predictions to the pixel level can improve model explainability and user confidence in the model.
A significant advantage of unsupervised AD approaches is eliminating the time-consuming and resource-intensive labeling process. Most Pixel-Level AD methods fall into two categories: reconstruction-based and feature embedding-based methods \cite{bao2023bmad} \cite{iad}. \\
\textbf{Reconstruction-based methods} use generative models to learn normal image reconstruction during training: large reconstruction errors during inference indicate anomalies. Approaches like AutoEncoders, GANs and Diffusion Models are prominent in this area.\\
\textbf{Feature Embedding-based methods} rely on data representations generated by pre-trained neural networks. These approaches can be further categorized into three categories as follows.\\
\textbf{Teacher-Student approaches} based on two networks (teacher and student) with knowledge distillation, where student and teacher feature map deviations indicate anomalies.\\
\textbf{Memory Bank approaches} that save features of normal images in a memory bank. Approaches like Padim, PatchCore, and CFA fall under this category.\\
\textbf{Normalizing Flow approaches} based on normalizing flow models to transform complex input data distributions into normal distributions, leveraging probability as a measure of normality.\\
These techniques offer diverse strategies for identifying anomalies while balancing interpretability and performance.
Most practical VAD applications must be deployed close to the environment of interest, requiring the processing of the models directly on edge devices.
Although the important realistic scenario, only \cite{paste} provides a benchmark for VAD algorithms on edge devices and proposes PaSTe: a new VAD model suited for edge deployment based on the STFPM architecture \cite{stfpm}.
However, the study is limited since the models are not able to handle dynamic environments where data can change over time and the model needs to adapt while remembering previous information.
For example, new objects could be introduced to the model over time.

\subsection{Continual Learning}
\label{subsec:continual_learning}

In traditional Machine Learning, models are trained on fixed datasets. However, real-world environments often present new data with different distributions from the training set. Continual Learning (CL) addresses this challenge by enabling models to adapt to new data without forgetting previous knowledge. Effective CL methods should minimize forgetting, have low memory usage, and be computationally efficient \cite{clad}.
CL techniques are generally categorized into three main approaches: rehearsal-based, regularization-based, and architecture-based. Rehearsal-based methods, such as Experience Replay \cite{expreplay}, store and reuse past data samples during training. Regularization-based approaches introduce constraints or penalties to retain knowledge of old tasks, often using parameter importance \cite{ewc} or distillation techniques \cite{lwf}. Architecture-based methods modify the model's structure to preserve prior knowledge \cite{progressive} \cite{fernando2017pathnet} \cite{mallya2018packnet}. 

Experience Replay is widely considered the most effective strategy to mitigate Catastrophic Forgetting \cite{yang2022benchmarkempiricalanalysisreplay} \cite{pellegrini2020latent} \cite{buzzega2020rethinkingexperiencereplaybag} \cite{kim2020imbalancedcontinuallearningpartitioning},  particularly for image classification. 
One of the first studies on Continual Learning for VAD has been performed in \cite{clad}, which shows how adaptations of VAD methods combined with a Replay strategy can get good performances with a quite long stream of tasks.
The CL scenario considered in \cite{clad} and in this work is shown in Fig. \ref{fig:CLAD_scheme}. 

In this work, we study the VAD algorithms for the first time in a Continual Learning scenario on the edge, where resources are limited.

%% file: sec/3_methodology.tex
\section{Methodology}
\label{sec:methodology}

\begin{figure*}[!th]
    \centering
    \begin{subfigure}{0.45\textwidth}
        \centering
        \includegraphics[width=\linewidth]{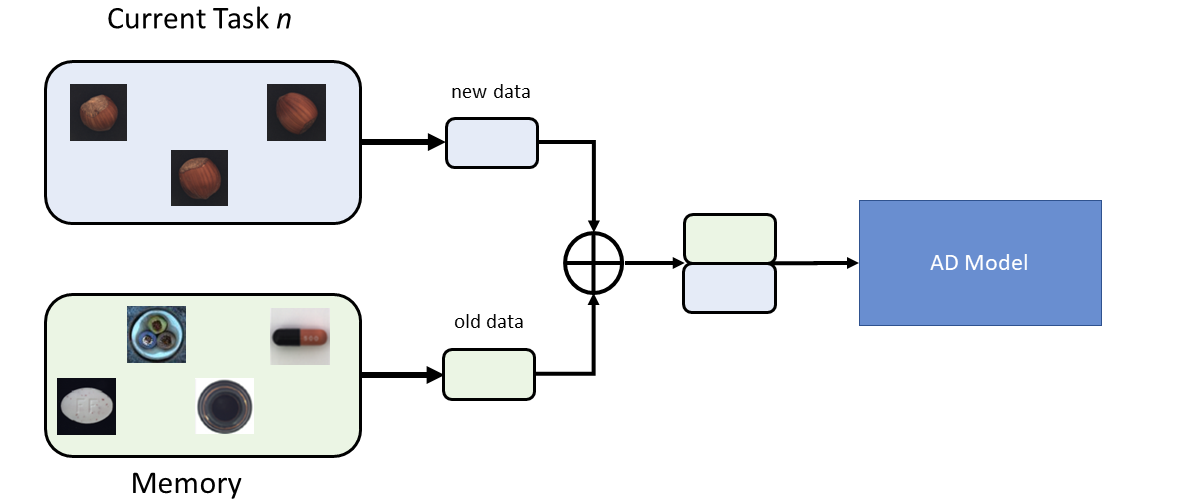}
        \caption{Replay}
        \label{fig:replay_approach}
    \end{subfigure}
    \hspace{0.05\textwidth}
    \begin{subfigure}{0.45\textwidth}
        \centering
        \includegraphics[width=\linewidth]{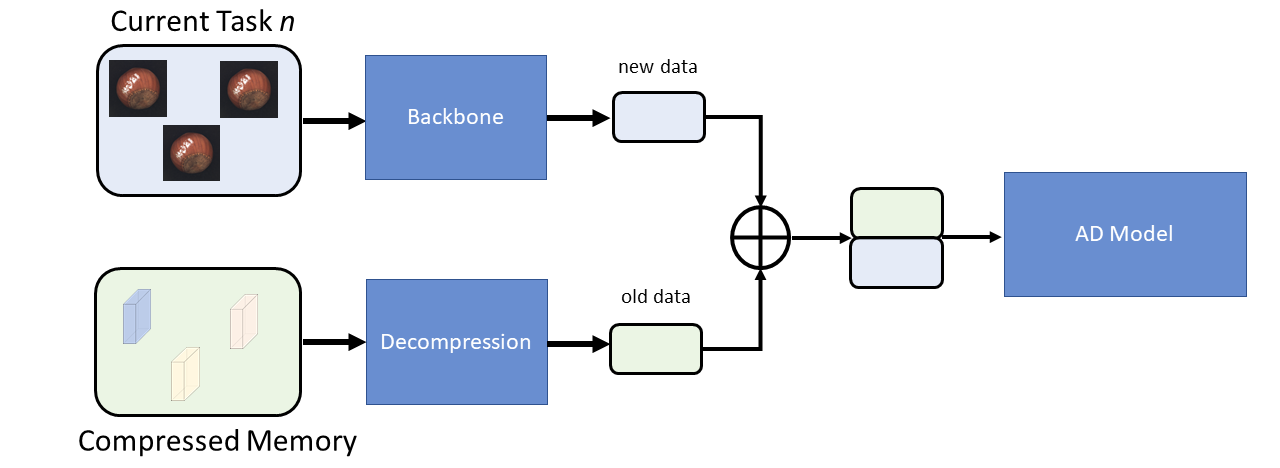}
        \caption{Compressed Replay}
        \label{fig:compressed_replay_approaches}
    \end{subfigure}
    \caption{On the left, we present the scheme of the Replay approach, where current data is combined with images from previous tasks, stored in the replay memory. On the right, we show the general framework for all Compressed Replay approaches, including Feature Replay, FQ Replay, and PQ Replay. In this approach, the replay memory stores either feature representations or compressed versions, which are then decompressed (when necessary) and then mixed with the representations of the images of the current task.}
    \label{fig:Replay_schemes}
\end{figure*}

\subsection{Continual Learning VAD on Edge}

Starting from the work done in \cite{paste} we test the STFPM model in the CL setting by adopting the tiny backbone MCUNet-IN3, which provides the best trade-off memory performance as reported in \cite{paste}. 
STFPM is evaluated considering Replay as CL method, Replay performs well for the VAD models as previously tested \cite{clad}. The same is done by considering the PaSTe architecture.
After that, further optimization of the Replay memory is performed by considering three Compressed Replay techniques as reported in Section \ref{subsec:compressed_replay}: Feature Replay, Feature Quantization (FQ) Replay, and Feature Product Quantization (FPQ) Replay.
The goal of Compressed Replay approaches aims to compress the replay memory to reduce the memory footprint.

\subsection{STFPM and PaSTe}
Student-Teacher Feature Pyramid Matching \cite{stfpm} is a feature-based VAD model that employs a teacher-student architecture for the task (see Fig. \ref{fig:STFPM_architecture}). Since the student replicates the features extracted by the teacher at given feature extraction layers on normal images, a difference between student and teacher features is an anomaly during inference. 
PaSTe is an optimized version of the STFPM, where the first layers of the network that are not used in the feature extraction are frozen and shared between the Student and the Teacher (see Fig. \ref{fig:Paste_architecture}). 
The backbone used for both STFPM and PaSTe in this work is the MCUNet-IN3 \cite{mcunet} pre-trained on the ImageNet dataset, which provides a good trade-off between memory and performance. 
For feature extraction, we consider the same layers reported in the work of VAD at the edge \cite{paste}.
Their results show that PaSTe, when compared to STFPM, decreases the inference time by 25\%, while reducing the training time by 33\% and peak RAM usage during training by 76\% (see Tab. \ref{Tab:summary_table_f1_part2}).

\subsection{Replay and Compressed Replay Approaches}
\label{subsec:compressed_replay}
\label{subsec:replay_approach}

The Replay approach is one of the most effective CL techniques to mitigate Catastrophic Forgetting \cite{yang2022benchmarkempiricalanalysisreplay} \cite{pellegrini2020latent} \cite{buzzega2020rethinkingexperiencereplaybag} and ca be described as in Fig. \ref{fig:replay_approach}. 
When a new task is introduced to the model, the model stores a small portion of samples into a storage or replay memory.
When the model is trained on a new task, to maintain the previously acquired knowledge, a batch of data is extracted from the replay memory and mixed with a batch of data from the current task.
This mix is then fed to the model to learn the new task while remembering previously seen tasks.

Since we are working with tiny devices, the additional memory needed for the Replay mechanism must be optimized and lowered.
As reported in \cite{clad}, the more old training samples are stored in the replay memory, the more catastrophic forgetting is reduced. 
Therefore, we decided to consider a series of Compressed Replay techniques that aim to reduce the memory footprint (MBs) while keeping the same number of samples (see Fig. \ref{fig:compressed_replay_approaches}).

We decided to optimize the replay memory for PaSTe by leveraging its particular architecture, where a portion of layers are shared between the teacher and the student (see Fig. \ref{fig:Paste_architecture}). 
Since these layers are shared and frozen, the trainable parameters of the network start after an intermediate layer, so the saved samples for Replay can be forwarded from there. 
Therefore, when referring to \textbf{Feature Replay}, it means there is no need to save the images in the replay memory but instead a latent representation of them (more precisely, a feature map extracted at an intermediate layer), which occupies less memory.

Therefore,  the storage in Compressed Replay approaches like Feature Replay is composed by representations (e.g., intermediate features) of the samples and can include a step of decompression when necessary (see Fig. \ref{fig:compressed_replay_approaches}).
From this point, further optimizations can be done on the representations stored in the replay memory. Precisely, the feature maps extracted can be compressed more by using feature quantization and product quantization. Then, during the training, the compressed representations can be decompressed.

\textbf{Feature quantization} is a commonly used technique for minimizing memory consumption, especially in resource-constrained environments. In this work, 8-bit quantization was applied to the feature maps stored in memory to reduce storage requirements significantly. 
Intermediate feature maps in PyTorch use 32-bit floating-point values, where each value occupies 4 bytes. Converting these to 8-bit integers through quantization reduces the memory footprint by a factor of four, resulting in more efficient memory usage.
Quantization was implemented by determining the minimum and maximum values of each
feature map and calculating a quantization scale based on the range. Each value in the feature map was scaled, rounded, and clamped within a range of 0 to 255, representing the 256 levels of 8-bit quantization. The quantization process also stores the minimum value and scaling factor for each feature map to ensure the feature maps can be accurately reconstructed. The scaling factor adjusts the range of the integers, while the minimum value shifts the range to align with the original feature map’s distribution. This enabled the reconstruction of the feature maps to approximate their original representation.

\textbf{Product Quantization (PQ)} is a powerful technique for reducing the memory footprint and computational complexity of high-dimensional vector representations \cite{pq}. The core idea of PQ is to decompose a high-dimensional vector space into a Cartesian product of lower-dimensional subspaces, each of which is quantized independently using Vector Quantization (VQ) techniques, such as k-means clustering. During this process, a vector $x \in R^d$ is split into $m$ sub-vectors, and each sub-vector is quantized by assigning it to the nearest centroid from a pre-learned codebook (set of centroids), resulting in a compact representation through centroid indices. PQ is particularly effective in large-scale applications such as image retrieval \cite{ir_pq} and natural language processing (NLP) \cite{nlp_pq}, where balancing speed, memory efficiency, and accuracy is critical. The hyperparameters of the PQ algorithm are
\begin{itemize}
    \item $num\_subvectors$ : number of subvectors to divide the initial vectors.
    \item $n\_bit\_subvector$ : number of bits for representing a given subvector. This is related to the number of centroids used in the clustering algorithm, which is $2^{n\_bit\_subvector}$.
\end{itemize}
By varying these parameters, it is possible to reduce the vectors dimensions in a more or less aggressive way.

\begin{table}[h!]
\centering
\caption{Comparison of STFPM and PaSTe for the MCUNet backbone as reported in \cite{paste} for the static scenario, where a different model is trained for each object.
While memory improvements are modest, there are significant gains in inference, training computation, and training memory, with a slight change in AD performance.
This suggests that using the Paste method could be essential for continual learning on edge devices with limited resources.
Here, $\Delta$ represents the improvement in percentual of PaSTe compared to STFPM.
}
\label{Tab:summary_table_f1_part2}
\begin{tabular}{lccc}
\toprule
 & \textbf{STFPM} & \textbf{PaSTe} & \textbf{ $\Delta$ [\%]} \\
\midrule
\textbf{Memory [MB]}         & 1.76   & 1.68   & 4.5   \\
\textbf{Inference [MAC]}     & 224.3M & 156.9M & 30.1  \\
\textbf{Training [MAC]}      & 146.85M& 91.12M & 37.95 \\
\textbf{AD Performance [FI]} & 0.52   & 0.53   & 0.09  \\
\bottomrule
\end{tabular}
\end{table}

%% file: sec/4_expsettings.tex
\section{Experimental Setting}
\label{sec:experimental_setting}

In the following part, we delineate the experimental setup employed in our work. The considered CL scenario is the same as that considered in \cite{clad} and visible in Fig. \ref{fig:CLAD_scheme}. Following this, we explain the metrics under consideration, which are shown in Tab. \ref{tab:summarytable}.
Lastly, we describe the CL strategies evaluated. 

\begin{figure}[thbp]
  \centering
  \includegraphics[width=0.6\linewidth]{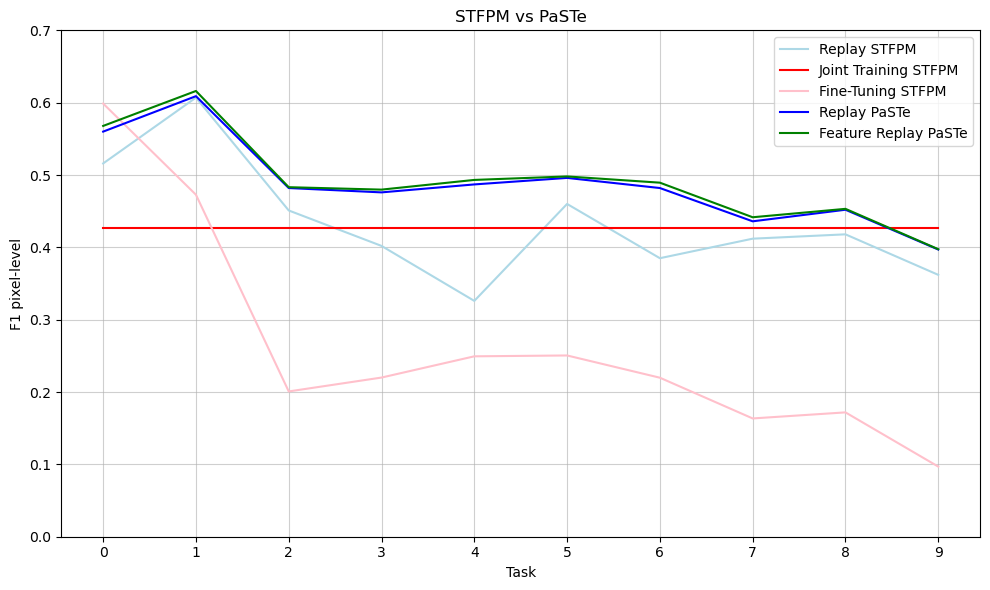}
   \caption{Behavior of the f1 pixel-level metric over time for STFPM and PaSTe. Each point represents the average performance on all the tasks seen so far. The considered strategies are the same as defined in \ref{clstrategies}}
   \label{Fig:stvspaste}
\end{figure}

\begin{figure}[thbp]
  \centering
  \includegraphics[width=0.6\linewidth]{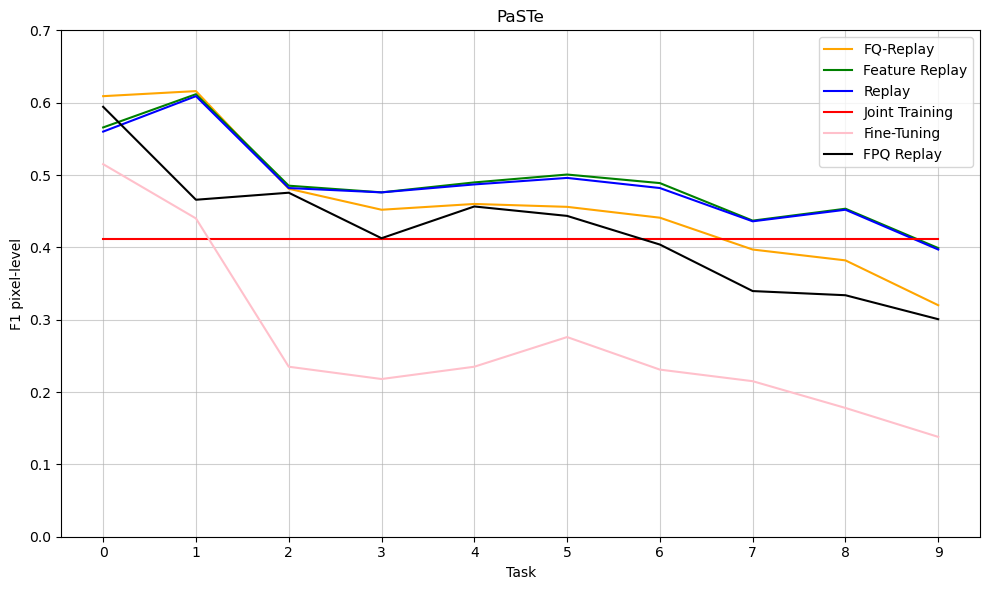}
   \caption{Behavior of the f1 pixel-level metric over time for PaSTe. Each point represents the average performance on all the tasks seen so far. The considered strategies are the same as defined in 4.2}
   \label{Fig:pastecl}
\end{figure}

\subsection {Dataset}
The MVTec Dataset \cite{mvtec} is used in our experiments. This dataset encompasses ten objects and five textures, making it suitable for assessing the robustness and generalization capabilities of various AD techniques. In the given CL scenario, we consider a sequence of ten tasks, where each task corresponds to a different object. Specifically, the evaluation includes the following objects: Bottle, Cable, Capsule, Hazelnut, Transistor, Metal Nut, Pill, Screw, Zipper, and Toothbrush.

\subsection {Continual Learning Strategies}
\label{clstrategies}

During the experiments, six CL approaches have been considered: 
\begin{enumerate}
    \item \textbf{Joint Training (JT)}: The model is trained with all the tasks simultaneously and is thus not affected by catastrophic forgetting. The performance obtained with this configuration is considered an upper bound for all the CL strategies. 
    \item \textbf{Fine-Tuning (FT)}: The model is trained by considering the data only from the current task without any countermeasure to catastrophic forgetting. The performance obtained with this configuration is considered a lower bound for all the CL strategies. 
    \item \textbf{Replay}: the model is trained by considering the data from the current task and a portion of the data from the previous tasks. In our experiment, the Replay memory size is 100, which means that the memory contains at most 10\% of the entire dataset.
    \item \textbf{Compressed Replay}: The model is trained as in the Replay strategy, but as stated before, we considered: 
    \begin{itemize}
        \item \textbf{Feature Replay:} In the memory, only the feature maps are stored.
        \item \textbf{Feature Quantization Replay (FQ Replay):} In the memory, the quantized feature maps are stored.
        \item \textbf{Product Quantization Feature Replay (FPQ Replay):} In the memory, the feature maps compressed with PQ are stored.
    \end{itemize}
\end{enumerate}

\subsection {Metrics}

In this section, all the VAD and CL metrics considered in the experiments are reported in Tab. \ref{tab:summarytable}.

\subsection {Anomaly Detection Metrics}

The performance of Anomaly Detection techniques using the MVTec dataset is typically evaluated using a range of metrics. A key distinction in evaluation approaches lies in the level of analysis, which can be conducted at either the image or pixel level. To provide a comprehensive assessment, we employ two primary metrics for image and pixel-level evaluations: the ROC and the F1-score.
In addition to these metrics, we also utilize the Per-region-overlap (PRO) metric for pixel-level evaluation. The PRO metric ensures that ground-truth regions are weighted equally, regardless of their size, thereby mitigating the limitations of simplistic per-pixel metrics \cite{yang2020improving}.

\subsection {Continual Learning Metrics}

The previous AD metrics are evaluated within the CL scenario. For each metric listed, the final value is averaged across all the tasks after the model is trained on the last task, as standard practice in the CL evaluations. 
For example, after the model was trained sequentially on all ten objects, the final performance is calculated by averaging the performance on all the objects.
The percentage of forgetting indicates the average forgetting on all the tasks and is defined as in \cite{clad}. A higher value of forgetting indicates it is more difficult to maintain previous knowledge.

To provide a more comprehensive evaluation, we also calculate the relative performance gap between each Continual Learning approach and the Joint Training, which is an upper bound for all CL methods. This comparison is crucial because a model may demonstrate minimal forgetting yet still achieve significantly lower overall performance than the Joint Training strategy. By highlighting this gap, we can better understand the trade-offs between knowledge retention and absolute performance in CL approaches.

As previously stated, in the CL framework, we are interested in updating and expanding model knowledge over time, avoiding forgetting, while at the same time, we want to obtain this goal with minimal computation and memory overhead \cite{de2021continual}. 
Hence, the values for the necessary memory are also provided in Tab. \ref{tab:summarytable}. For memory, we report the Architecture Memory and Additional Memory.
Architecture Memory represents the memory of the model and of the backbone, while Additional Memory refers to the memory necessary for the replay memory.

%% file: sec/5_results.tex
\section{Results}
\label{sec:results}

In this section, the results obtained with the various CL strategies explained in Sec. \ref{clstrategies} are discussed. Firstly, a comparison of the performances of STFPM and PaSTe with MCUNet-IN3 backbone with the Replay strategy is provided, and then a comprehensive analysis of the different Compressed Replay techniques adopted with PaSTe.

\begin{figure}[thbp]
  \centering
  \includegraphics[width=0.7\linewidth]{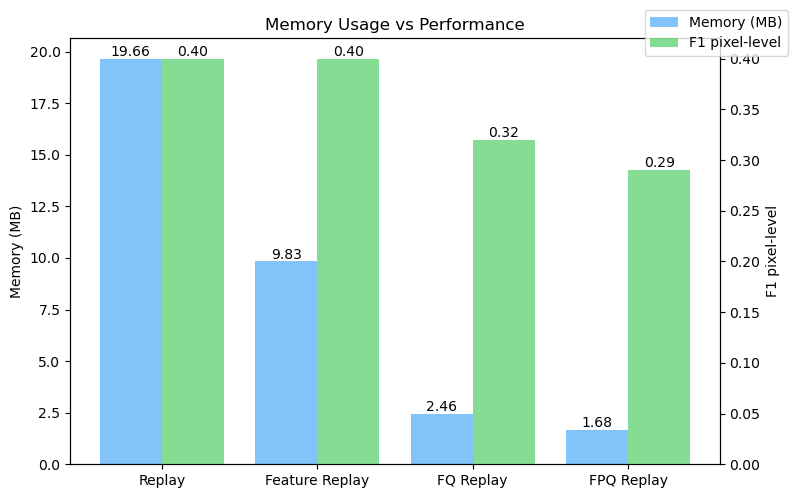}
   \caption{Memory - Performance trade-off between the different Replay strategies adopted with PaSTe}
   \label{Fig:memory}
\end{figure}

\subsection{Replay Approach}
In Fig. \ref{Fig:stvspaste}, a first comparison between STFPM and PaSTe models is shown, reporting the behavior of the F1 pixel level metric over time; each point represents the average performance on all the tasks seen so far. 
It is very interesting to note how PaSTe is performing better than STFPM in the Replay approach.
This is also confirmed by the values reported in Tab. \ref{tab:summarytable}, where the average forgetting value for PaSTe is lower than for STFPM. 

The motivation for this lower forgetting value can be found in the structure of the PaSTe model, where some layers are kept frozen and are shared between the Student and the Teacher. 
As a consequence, the initial layers generate more consistent feature representations across tasks than those obtained with a fully trainable Student-Teacher architecture. This property enables the Paste mechanism to adjust the weights of the trainable part less aggressively when transitioning between tasks, thereby reducing the extent of forgetting during model training.

\input{tables/summay_table}

\subsection{Compressed Replay Techniques}
As reported before, other optimizations have been considered to reduce the size of the replay memory in the PaSTe approach. 
A comprehensive overview of the trade-off between memory and performance can be depicted in Fig. \ref{Fig:pastecl} and Fig. \ref{Fig:memory}. 

A big improvement in the Replay memory size is achieved when switching from Replay to Feature Replay. Indeed, with the same number of saved samples in the memory (always 100), there is a reduction of 50\% of the Replay Memory size in terms of MBs, with no reduction in the F1 pixel level metric. Instead, when switching to Feature Quantization Replay, there is a reduction of 75\% with respect to the Feature Replay strategy with a decrease in the F1 pixel level metric (20\%). 

When adopting even more aggressive compression strategies, like PQ in the Feature Product Quantization Replay approach, there is another decrease of 68\% of the memory with another slight decrease in the F1 pixel level score (9.3\%). 
The results reported with FPQ Replay are obtained with hyperparameters: $num subvectors = 6$ \& $n bit subvector = 7$. 

Therefore, FPQ Replay achieves the highest compression ratio, providing a 12× reduction compared to classic Replay, followed by FQ Replay with an 8× reduction and Feature Replay with a 2× reduction.
However, based on the compression level, different AD performances are obtained, showing the trade-off between compression and performance (see Fig. \ref{Fig:memory}).

In particular, from these results, another big insight can be deduced. The Visual Anomaly Detection task is very precise, and a lot of precision in the extracted features is needed to perform the task well, at least for architectures like STFPM and Paste. By looking at the F1 pixel-level score reduction from Feature Replay to the FPQ Replay, it is easy to conclude that the quality of the feature maps that are saved in the Replay memory is crucial for the task, and altering them with a quantization process or more aggressively with Product Quantization, can lead to a decrease in the performances because it changes the feature distribution over which the Student and Teacher architectures must match their behavior.

%% file: tables/summay_table.tex
\begin{table*}[!th]
\caption{Table shows results for both STFPM and Paste model, the names of the CL approaches are the same as indicated in Sec. \ref{clstrategies}. Here are shown all the metrics for image-level and pixel-level. Moreover, additional metrics to consider for edge devices are reported, such as Architecture memory and Additional memory. Average forgetting is measured with respect to the f1 pixel metric.  }
\label{tab:results_complete}
\resizebox{1.0\textwidth}{!}{
\begin{tabular}{ll|cccc|ccccc}
\hline
\multicolumn{2}{c|}{Model}              & \multicolumn{3}{c|}{\textbf{STFPM}}                                                  & \multicolumn{6}{c|}{\textbf{PaSTe}}                                                                                                                                                \\ \hline
\multicolumn{2}{c|}{CL Approach}        & \multicolumn{1}{c|}{JT} & \multicolumn{1}{c|}{FT} & \multicolumn{1}{c|}{Replay} & JT & \multicolumn{1}{c|}{FT} & \multicolumn{1}{c|}{Replay} & \multicolumn{1}{c|}{\begin{tabular}[c]{@{}c@{}}Feature\\ Replay\end{tabular}} & \multicolumn{1}{c|}{FQ Replay} & PQ Replay \\ \hline
\multirow{2}{*}{\textbf{Image}}      & AUC ROC  & \multicolumn{1}{c|}{0.77}   & \multicolumn{1}{c|}{0.50}    & \multicolumn{1}{c|}{0.80}  & \multicolumn{1}{c|}{0.82} & \multicolumn{1}{c|}{0.54}   & \multicolumn{1}{c|}{0.80}   & \multicolumn{1}{c|}{0.82}   & \multicolumn{1}{c|}{0.69}  & 0.49  \\
                                      & $f_1$   & \multicolumn{1}{c|}{0.88}   & \multicolumn{1}{c|}{0.82}   & \multicolumn{1}{c|}{0.88}  & \multicolumn{1}{c|}{0.89} & \multicolumn{1}{c|}{0.83}   & \multicolumn{1}{c|}{0.87}   & \multicolumn{1}{c|}{0.89}  & \multicolumn{1}{c|}{0.86}      & 0.81      \\ \hline
\multirow{4}{*}{\textbf{Pixel}}      & AUC ROC & \multicolumn{1}{c|}{0.92}    & \multicolumn{1}{c|}{0.59}   & \multicolumn{1}{c|}{0.90}   & \multicolumn{1}{c|}{0.92} & \multicolumn{1}{c|}{0.74}   & \multicolumn{1}{c|}{0.90}   & \multicolumn{1}{c|}{0.91}   & \multicolumn{1}{c|}{0.88}      & 0.65      \\
                                      & $f_1$   & \multicolumn{1}{c|}{0.42}   & \multicolumn{1}{c|}{0.09}   & \multicolumn{1}{c|}{0.36}  & \multicolumn{1}{c|}{0.41} & \multicolumn{1}{c|}{0.15}   & \multicolumn{1}{c|}{0.40}   & \multicolumn{1}{c|}{0.40}  & \multicolumn{1}{c|}{0.32}      & 0.29      \\
                                      & PR AUC  & \multicolumn{1}{c|}{0.38}   & \multicolumn{1}{c|}{0.04}   & \multicolumn{1}{c|}{0.32}  & \multicolumn{1}{c|}{0.37} & \multicolumn{1}{c|}{0.10}   & \multicolumn{1}{c|}{0.36}   & \multicolumn{1}{c|}{0.36}  & \multicolumn{1}{c|}{0.26}      & 0.30      \\
                                      & AU PRO  & \multicolumn{1}{c|}{0.81}   & \multicolumn{1}{c|}{0.32}   & \multicolumn{1}{c|}{0.70}  & \multicolumn{1}{c|}{0.76} & \multicolumn{1}{c|}{0.44}   & \multicolumn{1}{c|}{0.74}   & \multicolumn{1}{c|}{0.76}  & \multicolumn{1}{c|}{0.70}      & 0.34      \\ \hline
\textbf{Architecture memory {[}MB{]}} &         & \multicolumn{1}{c|}{1.76}   & \multicolumn{1}{c|}{1.76}   & \multicolumn{1}{c|}{1.76}  & \multicolumn{1}{c|}{1.68} & \multicolumn{1}{c|}{1.68}   & \multicolumn{1}{c|}{1.68}   & \multicolumn{1}{c|}{1.68}  & \multicolumn{1}{c|}{1.68}      & 1.68      \\
\textbf{Additional memory {[}MB{]}}   &         & \multicolumn{1}{c|}{0}   & \multicolumn{1}{c|}{0}   & \multicolumn{1}{c|}{19.66} & \multicolumn{1}{c|}{0} & \multicolumn{1}{c|}{0}   & \multicolumn{1}{c|}{19.66}  & \multicolumn{1}{c|}{9.83}  & \multicolumn{1}{c|}{2.46}      & 1.68      \\
\textbf{Average forgetting {[}\%{]}}  &         & \multicolumn{1}{c|}{/}   & \multicolumn{1}{c|}{0.84}   & \multicolumn{1}{c|}{0.20}  & \multicolumn{1}{c|}{/} & \multicolumn{1}{c|}{0.76}   & \multicolumn{1}{c|}{0.19}   & \multicolumn{1}{c|}{0.19}  & \multicolumn{1}{c|}{0.28}      & 0.50      \\ \hline
\end{tabular}
}
\label{tab:summarytable}
\end{table*}

%% file: sec/6_conclusions.tex
\section{Conclusion}
\label{sec:conclusion}

In this work, we addressed the challenge of Continual Learning for Visual Anomaly Detection on edge devices, where memory and computational constraints pose significant challenges. 
Our study demonstrated the feasibility of deploying VAD models like STFPM on the edge while still being able to learn continuously from a data stream where new objects arrive over time.

Moreover, we tested the advantages of PaSTe, an optimized variant of STFPM, in a CL scenario for edge devices, improving efficiency in resource-limited environments.
Our results show that PaSTe not only reduces memory and computational requirements but also improves anomaly detection performance. Specifically, when combined with the Replay technique, PaSTe achieved a 10\% increase in pixel-level F1 score compared to STFPM while also reducing training time, inference time, and peak memory usage.
Moreover, we proved that Paste not only works well in the Continual Learning context, but it is also a more reasonable choice than STFPM when considering the edge application where the computation resources are limited.
Our results show that PaSTe is not only a lighter version of STPFM, but it also achieves superior anomaly detection performance, improving the f1 pixel performance by 10\% with the Replay technique.

Furthermore, the structure of PaSTe allows us to test it using a series of Compressed Replay techniques to further optimize memory consumption.
Our findings indicate that Feature Replay reduces memory usage by 50\% with no performance loss, while more aggressive compression strategies, such as Feature Quantization and Product Quantization, allow for a maximum memory reduction of 91.5\% but with performance degradation.
 
Eventually, our study demonstrates that Continual Learning can be effectively applied to VAD for edge deployment, enabling models to adapt incrementally to new tasks with minimal resource consumption. 
Future research directions include exploring alternative compression techniques to reduce further the replay memory or exploring other CL techniques like distillation-based approaches.